\documentclass[letterpaper]{article} 
\usepackage{aaai25}  
\usepackage{times}  
\usepackage{helvet}  
\usepackage{courier}  
\usepackage[hyphens]{url}  
\usepackage{graphicx} 
\usepackage{natbib}  
\usepackage{caption} 
\usepackage{algorithm}
\usepackage{algorithmic}
\usepackage{amsfonts}
\usepackage{mathrsfs}

\usepackage{multirow}
\usepackage{rotating}
\usepackage{booktabs}
\usepackage{subfigure}
\usepackage[utf8]{inputenc} 
\usepackage{nicefrac}       
\usepackage{microtype}      
\usepackage{xcolor}         
\usepackage[accsupp]{axessibility}

\usepackage{newfloat}
\usepackage{listings}
\DeclareCaptionStyle{ruled}{labelfont=normalfont,labelsep=colon,strut=off} 
\floatstyle{ruled}
\newfloat{listing}{tb}{lst}{}
\floatname{listing}{Listing}
\def\beqa{\begin{eqnarray}}
\def\eeqa{\end{eqnarray}}
\def\beqann{\begin{eqnarray*}}
\def\eeqann{\end{eqnarray*}}
\def\transpose{{^\top}}

\def\argmin{\mathop{\rm arg\, min}}

\newcommand{\red}[1]{{\textcolor{black}{#1}}}
\definecolor{mygreen}{rgb}{0.0, 0.5, 0.0}
\newcommand{\green}[1]{{\textbf{\textcolor{black}{#1}}}}

\title{Can Generative Models Improve Self-Supervised Representation Learning?}
\author {
    Sana Ayromlou\textsuperscript{\rm 1},
    Vahid Reza Khazaie\textsuperscript{\rm 1},
    Fereshteh Forghani\textsuperscript{\rm 2}\thanks{Work done during an internship at Vector Institute.},
    Arash Afkanpour\textsuperscript{\rm 1}
}
\affiliations {
    \textsuperscript{\rm 1}Vector Institute\\
    \textsuperscript{\rm 2}York University\\
    sana.ayromlou@vectorinstitute.ai \\
    vahidreza.khazaie@vectorinstitute.ai \\
    forghani@yorku.ca \\
    arash.afkanpour@vectorinstitute.ai
}

\begin{document}

\maketitle

\begin{abstract}
The rapid advancement in self-supervised representation learning has highlighted its potential to leverage unlabeled data for learning rich visual representations. However, the existing techniques, particularly those employing different augmentations of the same image, often rely on a limited set of simple transformations that cannot fully capture variations in the real world. This constrains the diversity and quality of samples, which leads to sub-optimal representations. In this paper, we introduce a framework that enriches the self-supervised learning (SSL) paradigm by utilizing generative models to produce semantically consistent image augmentations. By directly conditioning generative models on a source image, our method enables the generation of diverse augmentations while maintaining the semantics of the source image, thus offering a richer set of data for SSL. Our extensive experimental results on various joint-embedding SSL techniques demonstrate that our framework significantly enhances the quality of learned visual representations by up to 10\% Top-1 accuracy in downstream tasks. This research demonstrates that incorporating generative models into the joint-embedding SSL workflow opens new avenues for exploring the potential of synthetic data. This development paves the way for more robust and versatile representation learning techniques.
\end{abstract}

%

\section{Introduction}
\label{sec:intro}
\begin{figure*}[tb]
  \centering
  \includegraphics[width=0.73\textwidth]{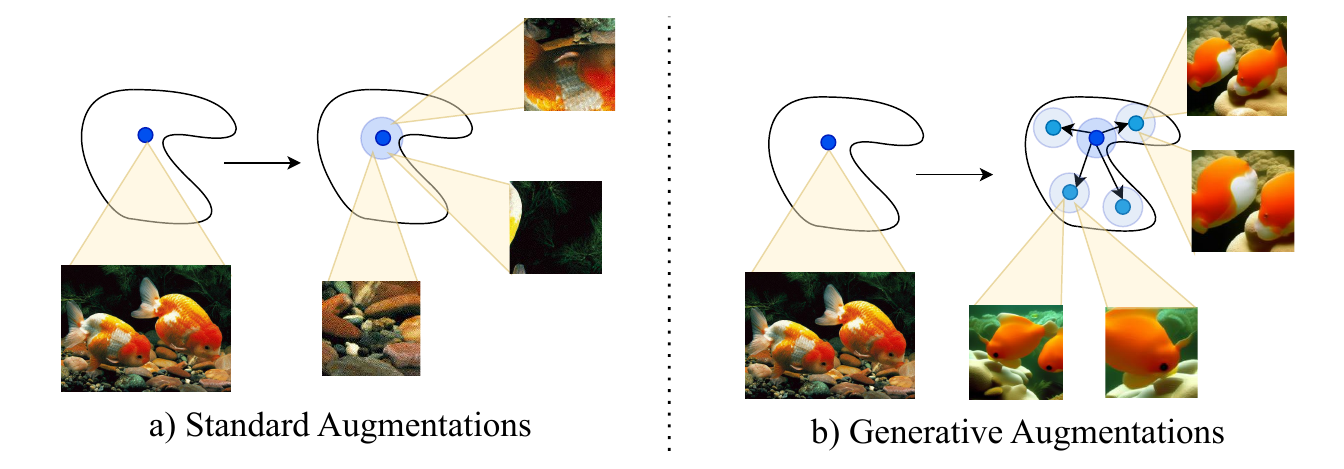}
  \caption{Generative augmentations produce a more diverse set of images with similar semantics. \textbf{a)} The standard SSL augmentations offer limited diversity for effective representation learning. \textbf{b)} By generating instance-conditioned samples, and then applying the standard augmentations on top, we add more diversity in training data, leading to better representations.}
  \label{fig:augmentations}
\end{figure*}

Self-supervised learning (SSL) is a machine learning approach where methods leverage large amounts of unlabeled data for representation learning. In the absence of any prior knowledge, most SSL methods assume that each data point is semantically different from other examples in a dataset. While this assumption can help devise a self-supervised representation learning algorithm, it is not sufficient to guarantee the generalization of representations to new tasks. As a result, \emph{joint-embedding} SSL methods (e.g., \citet{chen2020simple, he2020momentum, grill2020bootstrap}) utilize augmentation techniques to generate two or more views of the same image and map them to nearby points in the representation space. These augmentations typically consist of a series of transformations, such as random crop, color jitter, or Gaussian blur, as shown in Fig.~\ref{fig:augmentations}(a), which are used to define \emph{positive labels} for the SSL task. These augmentations determine pixel-space changes that should result in similar embedding-space representations. Since these transformations specify what representations learn, a natural question is whether additional transformations improve the generalization and robustness of representations.

While some previous works have studied the impact of augmentations in representation learning (e.g., \citet{chen2020simple}, \citet{caron2020unsupervised}), until recently, this area has remained largely under-explored. More recently, new image transformations have been introduced in the context of self-supervised representation learning. For example, \citet{mansfield2023random} introduced random field transformations by applying local affine and color transformations whose parameters are specified by Gaussian random fields \citep{adler2007random}. \citet{rojas2023sassl} proposed style transfer as a form of transformation for SSL. While these works show that new transformations can improve representation learning, their transformations remain limited to certain predefined forms.

An important feature of SSL transformations is that while they transform an image in pixel space, they do not significantly modify its semantics. Guided by this view, we propose the notion of a \emph{generalized transformation}, which satisfies these conditions: (1) the transformed image should be semantically similar to the original image and (2) it should not make modifications that render the image unrealistic, which often happens by some of the standard transformations (e.g., color jitter). The question arises: given a source image, how can one generate transformed images that satisfy these conditions? For this task, we leverage instance-conditioned generative models. In particular, we study latent diffusion models \citep{rombach2022high} and conditional Generative Adversarial Networks (GAN) \citep{casanova2021instance} that generate an output that is semantically similar to a source image. Our extensive empirical study in in-distribution and out-of-distribution cases demonstrates the effectiveness of our method for representation learning. The code to reproduce our empirical study is available at \url{https://github.com/VectorInstitute/GenerativeSSL}.
The key contributions of our work are as follows:
\begin{itemize}
    \item We propose to leverage instance-conditioned generative models for generating semantically similar augmentations in joint-embedding SSL, advancing beyond the standard augmentation techniques.
    \item We perform a rigorous and extensive empirical study of the instance-conditioned generative augmentation, demonstrating its effectiveness for learning representations with high generalization across in-distribution and out-of-distribution tasks.
    \item By relying on instance-conditioned image generation, we eliminate the need for text-based image generation, as suggested in \citet{tian2024stablerep}. This enables the application of our method to datasets without a textual description, resulting in a more versatile and effective strategy for SSL training.
\end{itemize}

\section{Background}
\label{sec:background}

\subsection{Self-supervised learning}
Self-supervised learning (SSL) leverages unlabeled data to create rich and generalizable representations. This approach offers numerous advantages due to its ability to train on extensive unlabeled data \citep{balestriero2023cookbook}. 
A line of SSL methods, known as \textit{joint-embedding} methods, encourages two views of the same image, formed by augmentations such as cropping or color jitter, to be mapped to similar representations. This approach enables encoders to learn to differentiate the representation of each image, leading to decent linear separability.

These methods, based on their objective functions or architectures, can be categorized as follows \citep{balestriero2023cookbook}: 1) \textit{Contrastive learning} methods employ the InfoNCE contrastive loss \citep{oord2018representation}, which in an unsupervised manner, pulls representations of different transformed versions of the same image together while pushing representations of different images apart. \textbf{SimCLR} \citep{chen2020simple} uses the examples in each batch to calculate the contrastive loss. \textbf{MoCo} \citep{he2020momentum} is another method that learns an encoder by matching an encoded query to a dictionary of encoded keys using contrastive loss. 2) \textit{Self-distillation} methods rely on a straightforward mechanism of feeding two different views to two encoders and mapping one to the other using a predictor~\citep{zhou2021ibot,caron2021emerging,oquab2023dinov2}. They prevent collapse by introducing asynchrony in updating the encoders. \textbf{BYOL}~\citep{grill2020bootstrap} first introduced self-distillation as a means to avoid collapse. \textbf{SimSiam}~\citep{chen2021exploring} replaces the BYOL moving average encoder with a stop-gradient mechanism. 3) \textit{Canonical correlation analysis} methods originate from the Canonical Correlation Framework (CCA)~\citep{hotelling1992relations, caron2018deep, bardes2021vicreg}. At a high level CCA aims to infer the relationship between variables by analyzing the covariance matrix. \textbf{Barlow Twins} \citep{zbontar2021barlow} uses cross-correlation of the embedding features to achieve self-supervised learning goals. In this paper, we evaluate the effectiveness of our proposed generative augmentation when integrated into the augmentation pipeline of the highlighted techniques (in bold).

\subsection{Generative methods}
\textit{Generative Adversarial Networks} (GAN) \citep{goodfellow2014generative} have significantly advanced the field of image generation. Recently, some work has been done on conditioning GANs over a feature vector~\citep{mangla2022data,casanova2021instance}. In this work, we chose Instance-Conditioned GANs (ICGAN)~\citep{casanova2021instance}, which extends this framework by conditioning the generation and discrimination processes on a specific image. ICGAN aims to model the data distribution as a mixture of local densities around each instance, utilizing representations of these instances as additional input to both the generator and the discriminator. This conditioning allows the model to generate images that are semantically similar to a given instance. This approach addresses the optimization difficulties and mode collapse issues prevalent in traditional GANs by ensuring a more focused generation process. This method can be leveraged to create semantically consistent augmentations for SSL.

\textit{Diffusion Models}, a class of likelihood-based generative models, have gained attention for their ability to generate high-quality images~\citep{dhariwal2021diffusion, ho2020denoising, nichol2021improved}. They work by gradually reducing noise from an image, with their training objective expressed as a re-weighted variational lower-bound~\citep{ho2020denoising}. Despite their ability to generate high-quality images, due to their incremental noise reduction process, these models have traditionally faced challenges of long training and inference times. While inference challenges can be mitigated with advanced sampling strategies~\citep{salimans2022progressive,song2020denoising} and hierarchical approaches~\citep{ho2022cascaded}, training on high-resolution image data remains computationally expensive. To address this, latent diffusion models (LDM) have been introduced~\citep{rombach2022high}. These models operate in a latent space with lower dimensionality than the input space, making training computationally cheaper and speeding up inference with almost no reduction in synthesis quality. They also introduced cross-attention layers into the model architecture, where various conditioning inputs, such as text, images, or other representations, can be applied to get a conditional output. The conditional LDM, $\epsilon_\theta$, which is usually a UNet \citep{ronneberger2015u}, and a domain-specific encoder, $\phi_\tau$, that provides the conditioning vector, are learned jointly by optimizing,
\beqann
    \theta^*, \tau^* = \argmin_{\theta, \tau} \hspace{1mm} \mathbb{E}_{\epsilon\sim\mathcal{N}(0,I),t,y} \left[ \| \epsilon - \epsilon_\theta (z_t, t, \phi_\tau(y)) \|_2^2 \right],
\eeqann
where $t$ is the timestep parameter, $z_t$ is the noisy latent vector at timestep $t$, and $y$ is the conditioning input (e.g. a text prompt). Given their ability to generate high-quality diverse samples, diffusion models stand out as a compelling option for our new generative augmentation.

\section{Related work}
\label{sec:related_work}
With the advance of generative models, the use of synthetic data has become a tool for training better models in various domains. In computer vision, synthetic data has been utilized to enhance model performance in tasks such as object detection~\citep{lin2023SynthdataOD}, semantic segmentation~\citep{chen2019learning,ros2016synthia}, and classification~\citep{azizi2023synthetic,sariyildiz2023fake}. In representation learning, synthetic data has been leveraged for multi-task learning~\citep{ren2018cross}, and manipulating latent data representation~\citep{liu2022palm,baradad2021learning, jahanian2021generative}. \citet{jahanian2021generative} investigated learning visual representations from synthetic data generated by black-box models, emphasizing the ability of latent space transformations to facilitate contrastive learning. In contrast to our work, they rely on latent space transformations to generate multiple views of the same semantic content. 
In addition, past works \citep{astolfi2023instance,zang2024diffaugenhanceunsupervisedcontrastive,li2025genview} present different methods for leveraging generative models for contrastive learning. \citet{astolfi2023instance} proposes an approach similar to ours for representation learning using ICGAN models. In contrast, our work presents a more rigorous study covering both ICGAN and Stable Diffusion as generative models.
The works of \citet{donahue2016adversarial,donahue2019large} introduce Bidirectional Generative Adversarial Networks (BiGAN) and BigBiGAN, which incorporate an inverse mapping mechanism within the GAN framework to enable both forward and inverse mappings between latent space and data. This allows for capturing semantic variations more effectively and making the representations valuable for auxiliary tasks.
More recently, StableRep \citep{tian2024stablerep} showed how visual representations learned from synthetic text-to-image data can be more effective than representations learned by using only real images in various downstream tasks. Unlike them, we do not replace a real dataset with a synthetic one. Instead, we leverage conditional generative models to enrich augmentations for self-supervised learning. Another distinction is that our method does not require text prompts and directly uses images as condition for the generative model.

\section{Generative self-supervised learning}
\label{sec:genssl}

Joint-embedding SSL methods create two or more views by applying a series of transformations to an image. More formally, given a batch of images, $\mathcal{B} = \{x_1, \ldots, x_N\}$, joint-embedding SSL methods learn an encoder, $f_{\theta}$, by optimizing a loss function, $\mathscr{L}$, over the batch, $\mathcal{B}$:   
\beqann
    && \theta^*,\psi^* = \argmin_{\theta,\psi} \\ && \mathbb{E}_{A_1,A_2,\mathcal{B}} \Big[ \mathscr{L} \Big( g_\psi(f_\theta(A_1(\mathcal{B}))), g_\psi(f'_\theta(A_2(\mathcal{B}))) \Big) \Big],
\eeqann
where $A_1$ and $A_2$ are augmentations applied to the images in $\mathcal{B}$ to create two views of each image, $g_{\psi}$ is a projection/prediction head applied to the output of the encoder before calculating the loss, and $f'$ is a variant of $f$ that, depending on the SSL algorithm, could be for instance, identical to $f$ (SimCLR), its exponential moving average (MoCo), or its frozen version (SimSiam).

The augmentations, $A_1$ and $A_2$, are a series of $K$ sequentially-applied transformations, $A_j =  T_K^{\gamma_K} \circ \ldots \circ T_1^{\gamma_1},$
where $T_i^{\gamma_i}$ is parameterized by $\gamma_i$. For example, a crop transformation could be parameterized by crop area, aspect ratio, etc. While parameterization of transformations increases the versatility of the views, they are often limited in producing semantically rich and diverse variations of the data. For example, previous SSL methods apply transformations such as random crop, color jitter, horizontal flip, grayscale (color removal), and Gaussian blur. Adding transformations that create more diversity in the output of the augmentation pipeline could potentially improve the generalization of learned representations.

\begin{figure*}[tb]
  \centering
  \includegraphics[width=0.75\textwidth]{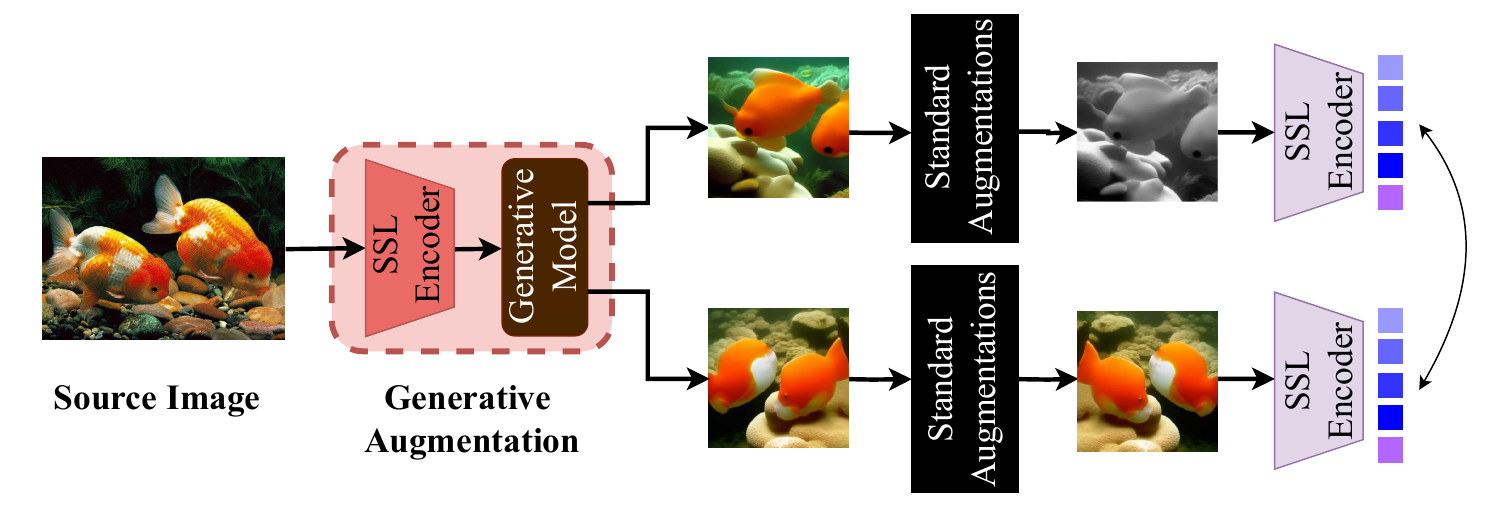}
  \caption{Our augmentation pipeline utilizes generative models, i.e., Stable Diffusion or ICGAN, conditioned on the source image representation, accompanied by the standard SSL augmentations. The components inside the \emph{Generative Augmentation} module, i.e. the pretrained SSL encoder and the generative model remain frozen throughout the SSL training process.}
  \label{fig:pipeline}
\end{figure*}

Due to their limited form, standard transformations might not adequately represent the intrinsic variability found in real-world data.
To overcome this limitation, we expand SSL transformations with a non-parametric transformation that is capable of enriching diversity of data. The only constraint is that to learn useful representations, a transformed image must remain semantically similar to its source image.

As illustrated in Fig.~\ref{fig:pipeline}, the foundation of this transformation lies in conditional generative models. These models, such as conditional latent diffusion models \citep{rombach2022high,bordes2021high}, instance-conditioned GANs \citep{casanova2021instance}, and conditional Variational Autoencoders (cVAEs) \citep{zhang2021conditional} enable the generation of images based on specific input data.
By conditioning the generation process on the input, the model ensures that the semantics of the original image is preserved to a large extent.
Moreover, the use of conditional generative models allows for a more nuanced augmentation process, where the generated samples closely follow the distribution of real images while exhibiting variations that enhance the diversity of the training data for self-supervised learning. Our new transformation is denoted by:
\beqann
T_0(x) =
\begin{cases} 
G(z;\phi(x)) & \text{if } \hspace{2mm} p \leq p_0 \\
x & \text{otherwise},
\end{cases}
\eeqann
where $G$ denotes a conditional generative model, $z \sim \mathcal{N}(0, I)$ is a noise vector, $\phi$ is a pretrained encoder, such as a CLIP image encoder \citep{radford2021learning}, that provides the condition vector for the generative model, $p \in [0, 1]$ is a random number, and $p_0$ is a parameter specifying the probability of applying the generative augmentation.

We emphasize that the new transformation and variants of it have been studied in a broader context of creating augmentations or views with generative models in previous work. See, for example, \citet{astolfi2023instance,zang2024diffaugenhanceunsupervisedcontrastive,li2025genview,jahanian2021generative}. In particular, \citet{astolfi2023instance} proposed a very similar augmentation using ICGAN models. What differentiates our work is that we perform a more extensive and rigorous study of this concept in the context of joint-embedding representation learning with two generative models across five SSL techniques and a suite of in-distribution and out-of-distribution tasks.


\section{Empirical study}
\label{sec:empirical_results}

\subsection{Setup}
\label{sec:setup}

Utilizing the Solo-learn library \citep{da2022solo}, we evaluate the proposed generative augmentation across five joint-embedding SSL techniques: SimCLR \citep{chen2020simple}, SimSiam \citep{chen2021exploring}, BYOL \citep{grill2020bootstrap}, Barlow Twins \citep{zbontar2021barlow}, and MoCo \citep{he2020momentum}. We pretrain a ResNet50 encoder using these SSL techniques for 100 epochs on the ImageNet training split. For evaluation, we follow the linear probing protocol of previous works by training a linear classifier on the output of the frozen encoder. Our downstream tasks are image classification on the ImageNet \citep{deng2009imagenet}, Food 101 \citep{bossard2014food}, Places 365 \citep{zhou2017places}, iNaturalist 2018 \citep{van2018inaturalist}, CIFAR 10, and CIFAR 100 \citep{krizhevsky2009learning} datasets. For all datasets, except for Places 365, a linear classifier is trained on the corresponding training split for 100 epochs. For Places 365, training is performed for 45 epochs. After training the linear classifier, we evaluate the classifier on the corresponding validation set.

In the following experiments, \emph{Baseline} refers to the model that uses standard augmentations to create the views. The sequence of transformations to create each view is as follows: (1) random crop with the relative crop area selected randomly from $[0.2, 1]$, (2) scale to 224$\times$224, (3) color jitter, (4) grayscale, (5) Gaussian blur, (6) horizontal flip. Each augmentation is applied stochastically with a probability value. More details are available in the Appendix.
For our method, we prepend the above sequence with a generative augmentation that takes a source image and, depending on the experiment, uses either Stable Diffusion, which is a conditional latent diffusion model, or ICGAN, to return a synthetic image as described in Section~\ref{sec:genssl}. To accelerate training, instead of generating synthetic images on-the-fly, we generate 10 images per each image in the ImageNet training set offline. Furthermore, with a given probability $p$ we apply the generative augmentation by selecting and loading one of these augmentations during SSL training. 

\subsection{Results}
\label{sec:results}

We frame our experiments to answer four research questions about our proposed generative augmentation technique.

\subsubsection*{RQ1: What probability of applying the generative augmentation achieves the best visual representations?}
Each image transformation in the standard augmentations is applied with a certain probability value tuned to achieve the best visual representations. While higher probability values create more diversity in augmentations, they could potentially hinder fidelity or create undesirable artifacts that result in sub-optimal representations.

\begin{figure}[thb]
    \centering
    \includegraphics[width=0.42\textwidth]{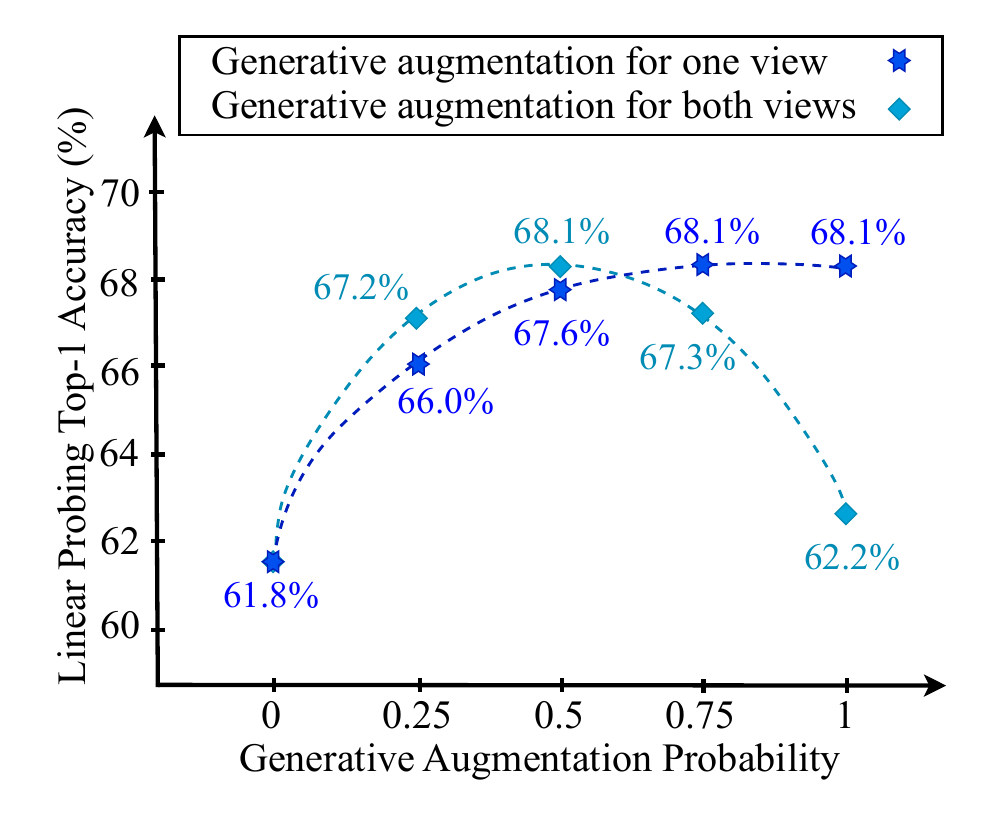}
    \caption{The effect of different probability values of applying the generative augmentation.}
    \label{fig:main_results_right}
\end{figure}

For this experiment we used Stable Diffusion for synthetic image generation and applied the new augmentation with probability $p \in \{0, 0.25, 0.5, 0.75, 1\}$. For each value of $p$, we trained a SimCLR encoder, and compared the encoders with linear probing on the ImageNet validation split. Figure~\ref{fig:main_results_right} shows the results. We ran two sets of experiments. In one set, we applied the generative augmentation only to one view (dark blue curve), while in the other set we apply this augmentation to both views (light blue curve). In all cases the standard augmentations are applied to both views. For the \emph{one} view, the results show a monotonic increase in the quality of the visual representations as $p$ increases with $p=0.75$ and $p=1$ achieving the best results. These results show the effectiveness of the new augmentation for representation learning. On the other hand, the best results are achieved with $p=0.5$ for \emph{both} view, as it strikes a balance between real and synthetic images in the views. However, increasing the probability of the generative augmentation beyond that will degrade performance as it likely creates excessive deviation from the original data. However, note that in this case $p=1$ still outperforms no generative augmentation ($p=0$), indicating the effectiveness of the new augmentation.

\subsubsection*{RQ2: Does the generative transformation improve representation learning consistently across different joint-embedding SSL techniques?}
\begin{figure*}[thb]
  \centering
  \includegraphics[width=0.65\textwidth]{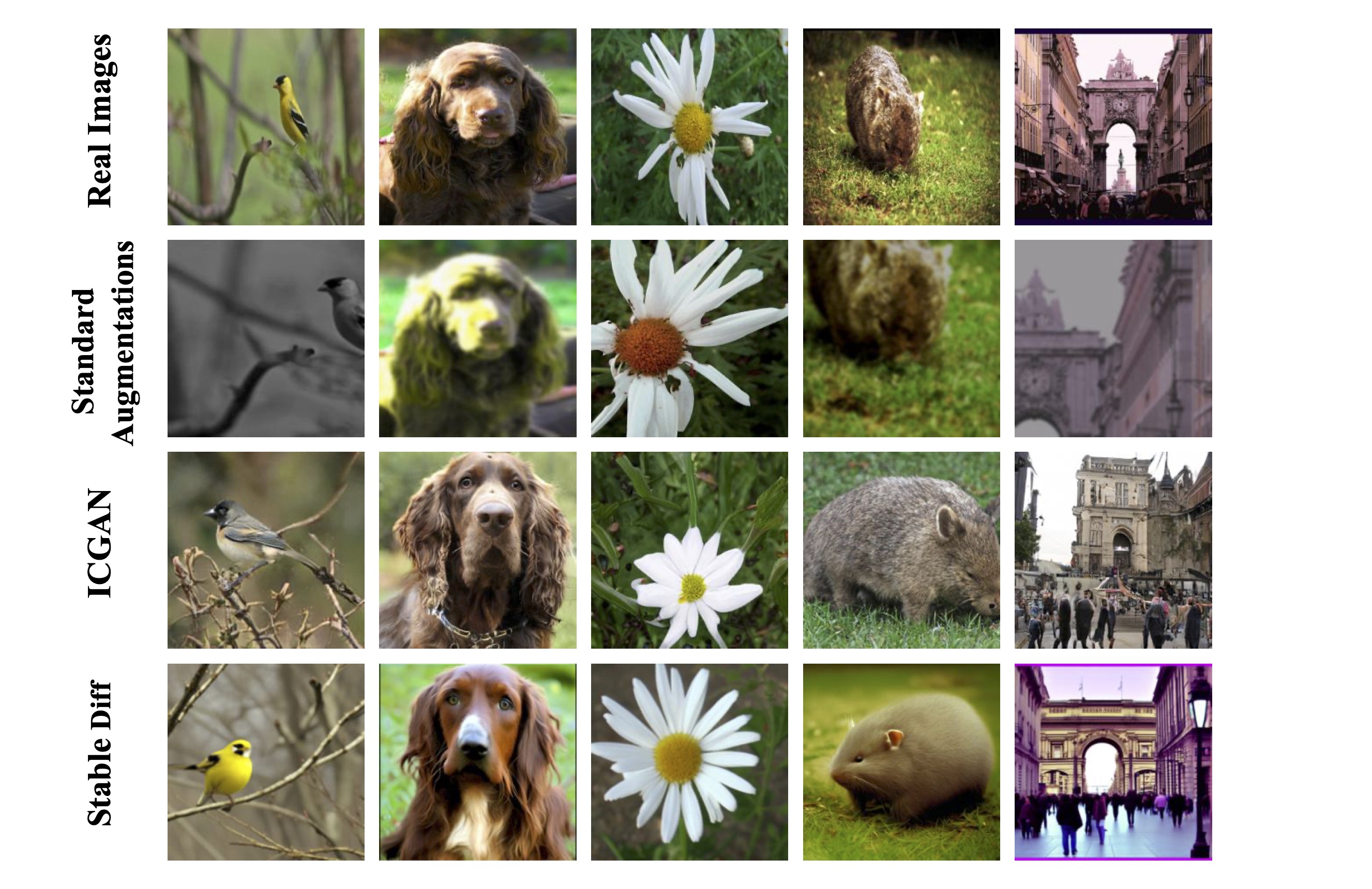}
  \caption{Examples of various augmentations. Compared to the standard augmentations (second row), instance-based generative augmentations can produce more diverse and realistic images that preserve the semantics of the original image.}
  \label{fig:example}
\end{figure*}

\begin{figure}[thb]
    \centering
    \includegraphics[width=0.42\textwidth]{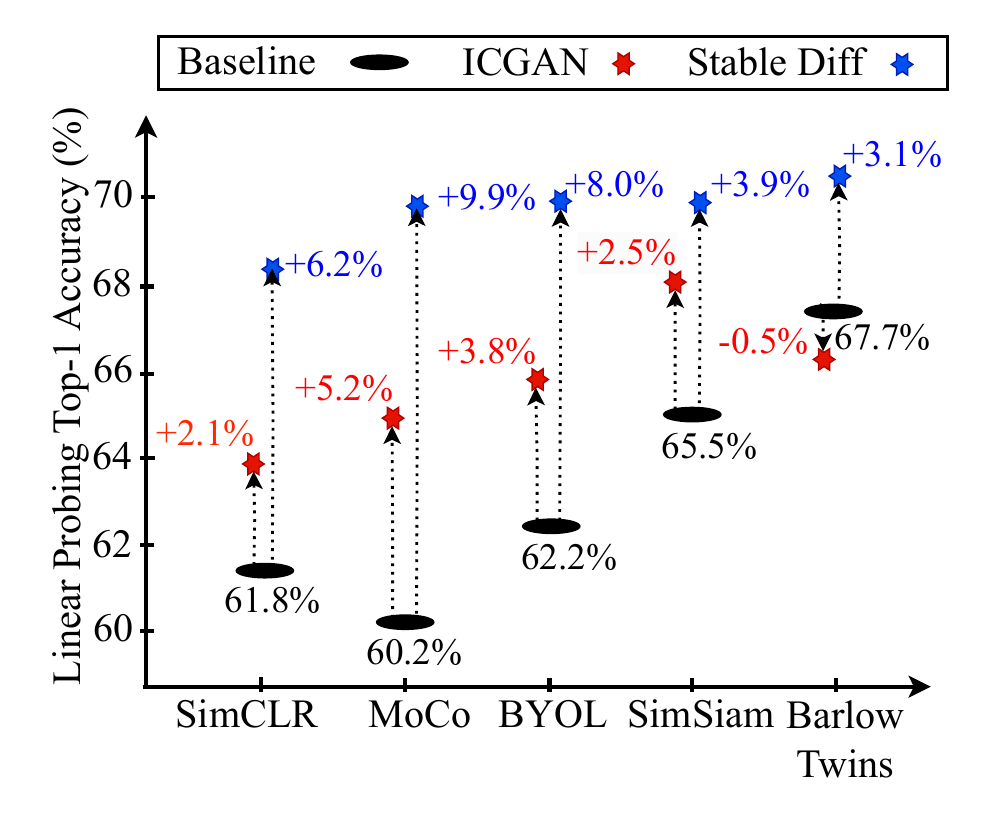}
    \caption{Top-1 accuracy improvement on ImageNet validation set obtained by the generative augmentations with Stable Diffusion and ICGAN across five SSL techniques. }
    \label{fig:main_results_left}
\end{figure}

We evaluate the effectiveness of the generative transformation across five SSL techniques. We compare an augmentation pipeline that includes the generative transformation obtained by either Stable Diffusion (Stable Diff) or ICGAN against the Baseline, which refers to pretraining the SSL encoder with only the standard augmentations. Based on the results depicted in RQ1, we apply the generative augmentation with probability $p=0.5$ to both views. Fig.~\ref{fig:main_results_left} shows the linear probing Top-1 accuracy for different SSL techniques. Additionally, Top-5 accuracy results are reported in the Appendix. Across all techniques, the addition of a generative transformation via Stable Diffusion consistently improves representation learning, which in turn enhances downstream classification accuracy between 3\% to 10\%. With the exception of Barlow Twins, we observe a similar trend with generative augmentations generated by ICGAN. However, the magnitude of improvement is smaller compared to Stable Diffusion. We conjecture that this is because the quality and diversity of images generated by Stable Diffusion better reflect the distribution of real images compared to those generated by ICGAN, as depicted by some examples in Fig.~\ref{fig:example}.
These factors are often determined by the size of the model and the volume of pretraining data. Stable Diffusion is a larger model and has been trained on LAION-400M which contains 400 million images \citep{rombach2022high}, while ICGAN was trained on ImageNet with 1.2 million images. 


To measure out-of-distribution performance of learned representations, we evaluate these models across other downstream classification tasks. Table~\ref{tab:ood_top1} shows the Top-1 accuracy results. In line with the above results on ImageNet, the Stable Diffusion generative transformation outperforms other baselines across all datasets. ICGAN, with the exception of a few cases, also outperforms the baseline, but we observe larger improvement with Stable Diffusion compared to ICGAN. These results, along with the in-distribution results, indicate that the generative transformation consistently improves representation learning across different SSL techniques.

\begin{table*}[t]
\small
\centering
\begin{tabular}{c|c|c|c|c|c|c}
\toprule
\multicolumn{1}{c}{} & & \textbf{Food101} & \textbf{CIFAR10} &\textbf{CIFAR100} & \textbf{Places365} & \textbf{iNaturalist2018}\\
\midrule
\multirow{4}{*}{\begin{turn}{-270}\textbf{SimCLR}\end{turn}} & Baseline & 58.03 & 61.74 &  36.88 & 46.11 & 20.17 \\
\cmidrule{2-7}
& ICGAN & 57.62 (\red{-0.41}) & 61.92 (\green{+0.18}) & 38.74 (\green{+2.06}) & 47.41 (\green{+1.67})& 20.01 (\red{-0.16})\\
\cmidrule{2-7}
& Stable Diff & 62.76 (\green{+4.73}) & 63.35 (\green{+1.61})  & 40.17 (\green{+3.29})  & 48.83 (\green{+2.62})& 24.13 (\green{+3.96})\\
\midrule
\midrule
\multirow{4}{*}{\begin{turn}{-270}\textbf{MoCo}\end{turn}} & Baseline & 54.50 & 58.17 & 34.28 & 45.70  & 16.52 \\
\cmidrule{2-7}
& ICGAN & 59.54 (\green{+5.04}) & 63.70 (\green{+5.53}) & 39.90 (\green{+4.62}) & 48.82 (\green{+3.12})& 20.99 (\green{+4.47})\\
\cmidrule{2-7}
& Stable Diff & 63.97 (\green{+9.47})& 65.14 (\green{+6.97})& 41.21 (\green{+6.93}) & 49.91 (\green{+4.21})& 26.29 (\green{+9.97})\\
\midrule
\midrule
\multirow{4}{*}{\begin{turn}{-270}\textbf{BYOL}\end{turn}} & Baseline & 54.14 & 55.34 &  28.70 & 46.93 & 6.91  \\
\cmidrule{2-7}
& ICGAN & 53.12 (\red{-0.98})& 58.76 (\green{+3.42})& 34.49 (\green{+5.79}) & 47.24 (\green{+0.31}) & 7.37 (\green{+0.46}) \\
\cmidrule{2-7}
& Stable Diff & 58.85 (\green{+4.71})& 61.90 (\green{+5.56})& 38.71 (\green{+10.01}) & 49.09 (\green{+2.16}) & 10.74 (\green{+3.83})\\
\midrule
\midrule
\multirow{4}{*}{\begin{turn}{-270}\textbf{SimSiam}\end{turn}} & Baseline & 60.71 & 61.93 & 37.69 & 48.78  & 22.07 \\
\cmidrule{2-7}
& ICGAN & 65.15 (\green{+4.44}) & 64.28 (\green{+0.35}) & 40.21 (\green{+2.52}) & 49.38 (\green{+0.60})& 27.07 (\green{+5.00})\\
\cmidrule{2-7}
& Stable Diff & 64.85 (\green{+4.14})& 63.88 (\green{+1.95})& 40.58 (\green{+2.89}) & 49.66 (\green{+0.88}) & 28.25 (\green{+6.18}) \\
\midrule
\midrule
\multirow{4}{*}{\begin{turn}{-270}\textbf{\shortstack{Barlow \\ Twins}}\end{turn}} & Baseline & 66.71 & 65.49 &  41.19 & 49.47 & 27.99 \\
\cmidrule{2-7}
& ICGAN  & 65.50 (\red{-1.21}) & 63.34 (\red{-2.15})& 41.95 (\green{+0.76}) & 48.64 (\red{-0.83}) & 26.19 (\red{-1.80}) \\
\cmidrule{2-7}
& Stable Diff & 69.97 (\green{+3.26}) & 65.20 (\red{-0.29})& 43.47 (\green{+2.28}) & 50.31 (\green{+0.84}) & 32.42 (\green{+4.43}) \\
\bottomrule
\end{tabular}
\caption{Top-1 accuracy ($\%$) results of linear probing on several datasets with different augmentation strategies. Numbers inside parentheses show the absolute percentage difference from the baseline, with bold numbers indicating positive improvements.}
\label{tab:ood_top1}
\end{table*}

\subsubsection*{RQ3: Considering that the new generative augmentation employs a conditional generative process dependent on a pretrained encoder, does self-supervised learning with this transformation produce the same latent-space data distribution as the pretrained encoder?}

This question is crucial because our augmentation technique relies on conditional image generation using a pretrained encoder for image conditioning (e.g., a pretrained CLIP encoder for Stable Diffusion). If the distributions of representation vectors are identical or very similar, the new data augmentation would offer little to no advantage for representation learning, as such representations are already accessible through the pretrained SSL encoder. Note that in this case we cannot rely on downstream accuracy for two reasons: (1) comparing downstream accuracy of two encoders provides little insight into the similarity of their representation spaces, (2) the CLIP encoder is trained on 400 million image-text pairs \citep{radford2021learning}, which is significantly larger than the ImageNet training split (1.2 million images) used for training our SSL encoders, rendering any downstream accuracy comparison unfair.
Instead, to compare representations of different encoders we employ two dissimilarity measures, i.e. Centered Kernel Alignment (CKA) \citep{kornblith2019similarity} and Orthogonal Procrustes Distance (OPD) \citep{ding2021grounding}, which have been used to compare representations between layers of a network or between different trained
models. Let $\mathcal{D} = \{ x_i \}_{i=1}^N$ be a dataset of $N$ examples, where each example $x_i \in \mathbb{R}^s$. Let $f_A: \mathbb{R}^s \rightarrow \mathbb{R}^p$ and $f_B: \mathbb{R}^s \rightarrow \mathbb{R}^q$ denote two encoders. Let $A \in \mathbb{R}^{p\times N}$ and $B \in \mathbb{R}^{q\times N}$ denote the representation matrices obtained by applying $f_A$ and $f_B$ to data points in $\mathcal{D}$ respectively. The dissimilarity measure based on CKA is defined by,
\beqann
    d_{\text{CKA}}(A,B) = 1 - \frac{\|AB\transpose\|_F^2}{\|AA\transpose\|_F \|BB\transpose\|_F}.
\eeqann
The OPD dissimilarity measure is the solution to the following optimization problem:
\beqann
    d_{\text{OPD}}(A,B) = \min_R \| B - RA \|_F^2, \hspace{5mm} \text{s.t.} \hspace{1mm} R\transpose R = \mathbb{I}.
\eeqann
Here, $\| . \|_F$ denotes Frobenius norm and $\mathbb{I}$ is the identity matrix.
Per standard practice, we normalize each matrix by first centering the features around the origin, then dividing by the Frobenius norm of the matrix before calculating these measures.

For this analysis, we measure the dissimilarity of representations between CLIP (the encoder used for conditioning Stable Diffusion) and SimCLR encoders trained with our generative transformation. We use the ImageNet validation set as data points to build the representation matrices. Since $d_{\text{CKA}}$ and $d_{\text{OPD}}$ are measured between two sets, we first require to establish a baseline for the dissimilarity value between two sets of similar representations so we can compare other values against it. For this purpose, we trained two SimCLR encoders with different weight initialization. While we acknowledge that the representation spaces of these models might not be identical, the dissimilarity value between them can be used as a baseline for comparing other values. Table~\ref{tab:dissimilarity} presents the dissimilarity values between different encoders.
\begin{table}[tb]
\small
\centering
\begin{tabular}{l|c|c}
\toprule
\textbf{Representation Pair} & \textbf{CKA} & \textbf{OPD} \\ \midrule
(SimCLR$_1$, SimCLR$_2$) & $0.395 \pm 0.006$ & $0.022 \pm 0.001$ \\
(SimCLR$_1$, CLIP) & $0.964 \pm 0.002$ & $0.580 \pm 0.002$ \\
(SimCLR$_2$, CLIP) & $0.952 \pm 0.002$ & $0.574 \pm 0.002 $ \\
\bottomrule
\end{tabular}
\caption{CKA and OPD dissimilarity values between different representation pairs. We report the mean and 95\% bootstrap confidence intervals over 100 runs.}
\label{tab:dissimilarity}
\end{table}
Dissimilarity values between the SimCLR encoders measured by both CKA and OPD are significantly smaller than the values between each SimCLR encoder and the CLIP encoder. These results demonstrate that the representations of SimCLR trained with the generative transformation are different from the representations of CLIP. In other words, representation learning with the proposed generative transformation does not trivially yield the same representation space of the pretrained encoder of the generative process.

\subsubsection*{RQ4: Can we replace the standard augmentations with the generative augmentation for self-supervised learning?}
Previous work on SSL augmentations has focused on enhancing the diversity of augmentations by adding new transformations to the pipeline. To the best of our knowledge, no attempt has been made to replace the standard augmentations with a new one, as it could significantly reduce the diversity of views and result in poor representations. Notably, \citet{tian2024stablerep} observed that contrastive learning with a limited number of synthetic images benefits from the standard augmentations, as they help reduce overfitting. We investigate whether the new augmentation can replace the standard augmentations and whether keeping the standard augmentations in the pipeline is still beneficial for visual representation learning.

For this study we trained SimCLR models with different augmentation strategies and compared their representations by linear probing on the ImageNet validation set.
Here, \emph{Baseline} refers to the standard augmentations (crop and resize, color jitter, grayscale, Gaussian blur, and horizontal flip), and \emph{Only Generative} refers to using only the new generative augmentation obtained from Stable Diffusion. We also included another strategy, namely \emph{Generative \& Random Crop} that applies the generative transformation, followed by the random crop transformation. We particularly chose random crop as \citet{chen2020simple} noted that this is the most effective augmentation among the standard ones. 
For completeness, we included \emph{Generative \& Standard} in the results, which applies the generative augmentation followed by the standard augmentations.
Table~\ref{tab:rq4} shows the linear probing Top-1 accuracy of these models. Applying the generative augmentation and excluding the standard augmentations results in 1.96\% improvement in accuracy. However, the results of \emph{Generative \& Random Crop} and \emph{Generative \& Standard} indicate that further improvement is gained by including the standard augmentations. These results demonstrate that the standard augmentations cannot be replaced by the new generative augmentation and their combination achieves the best result.

\begin{table}[tb]
\small
\centering
\begin{tabular}{l|l}
    \toprule
    \textbf{Augmentation Strategy}  & \textbf{Top-1 Accuracy} \\
    \midrule
    Baseline (only Standard) & 61.84 \\
    \midrule
    Only Generative & 63.80 (\green{+1.96}) \\
    \midrule
    Generative \& Random Crop & 67.15 (\green{+5.31}) \\
    \midrule
    Generative \& Standard & 68.11 (\green{+6.27}) \\
    \bottomrule
\end{tabular}
\caption{The effect of different augmentation strategies on representation learning, measured by downstream Top-1 accuracy on the ImageNet validation set. Numbers inside parentheses show the absolute difference (\%) from the baseline, with bold numbers indicating positive improvements..}
\label{tab:rq4}
\end{table}

\section{Conclusion}
\label{sec:conclusion}
Image augmentation is a crucial component of joint-embedding self-supervised learning. 
While many techniques rely on a limited set of transformations, recently, a few studies have proposed new parametric pixel-space transformations to enhance augmentations. In this paper, we studied a new generalized augmentation method that leverages instance-conditioned generative models to transform an image while preserving its semantics.
By relying on strong generative models this transformation is capable of producing diverse and realistic images.
Our empirical results demonstrate the effectiveness of the generative augmentations for self-supervised representation learning.
Our method currently relies on pretrained generative models. An interesting direction to expand this work is to co-train an instance-conditioned generative model and the SSL encoder simultaneously. While care must be taken to avoid a possible collapse to trivial solutions, this approach could potentially lead to enhanced generative models for the SSL task.

\section{Ethics statement}
Our method illustrates the potential for improving visual representation learning through the use of synthetic data. We recognize the risks of using synthetic data from pretrained generative models. Such models may have been trained on biased data, which could propagate or even amplify these biases in downstream applications. Data leakage is also possible if the generative model has not been trained with privacy-preserving methods. Our method, however, does not contribute to or exacerbate data leakage.

\bibliography{main}

\clearpage
\appendix
\section{Supplementary Material}

In this supplementary material, we provide a detailed overview of our method's implementation, including the exact process for applying various augmentations, as well as the settings used for both training and linear probing. We also present additional results that complement the experiments described in the main paper. Furthermore, we include samples of generated images using both Stable Diffusion and ICGAN. Finally, we provide information regarding the computational resources used in our experiments.

\section{Implementation details}
\label{sec:app_aug_config}

\subsection{Standard SSL augmentations}
We follow the default standard augmentation settings for each joint-embedding SSL algorithm provided in the solo-learn library~\citep{da2022solo}. These augmentations can be symmetric or asymmetric across the two views of these algorithms. The parameters of these symmetric and asymmetric standard augmentations are provided in Tables~\ref{tab:simclr_aug} and \ref{tab:others_aug}, respectively. Additionally, to perform ablation studies for \textbf{RQ4} in Section~\ref{sec:results}, we experiment with an extra random crop augmentation, where only crop and flip augmentations are applied as standard augmentations. 

\begin{table*}[hbt!]
\centering
\begin{tabular}{l|c|c|c | c }
\toprule
& \textbf{Parameter} & \textbf{\shortstack{SimCLR \\ Standard}} & \textbf{\shortstack{SimCLR \\ Random Crop}} & \textbf{\shortstack{MoCo \\ Standard}} \\
\midrule
\midrule
\multirow{2}{*}{Crop} & Min Scale  &  0.08   & 0.08 & 0.08   \\
\cmidrule{2-5}
&  Max Scale &   1.0 & 1.0 & 1.0  \\
\midrule
\midrule
\multirow{6}{*}{Color Jitter} & Prob    & 0.8   & 0.0 & 0.8\\
\cmidrule{2-5}
& Brightness  &  0.8  & -  &  0.4\\
\cmidrule{2-5}
& Contrast  &  0.8  & - &  0.4\\
\cmidrule{2-5}
& Saturation  &  0.8  & - &  0.4\\
\cmidrule{2-5}
& Hue  &  0.2   & - &  0.1\\
\midrule
\midrule
\multirow{1}{*}{Gray Scale} & Prob  &  0.2  & 0.0  &  0.2 \\
\midrule
\midrule
\multirow{1}{*}{Gaussian Blur} & Prob  &  0.5   & 0.0  &  0.5\\
\midrule
\midrule
\multirow{1}{*}{Solarization} & Prob  &  0.0   & 0.0 &  0.0 \\
\midrule
\midrule
\multirow{1}{*}{Horizontal Flip} & Prob  &  0.5  & 0.5 & 0.5\\
\bottomrule
\end{tabular}
\caption{The augmentation parameter values for SimCLR and MoCo are symmetric across both views.}
\label{tab:simclr_aug}
\end{table*}

\begin{table*}[hbt!]
\centering
\begin{tabular}{l | c | c | c | c }
\toprule
 & \textbf{Parameter} & \textbf{\shortstack{BYOL \\ Standard} } & \textbf{\shortstack{SimSiam \\ Standard} }& \textbf{\shortstack{Barlow Twins \\ Standard} } \\
\midrule
\midrule
\multirow{2}{*}{Crop} & Min Scale   & 0.08   & 0.08   & 0.08   \\
\cmidrule{2-5}
&  Max Scale  & 1.0  & 1.0 & 1.0 \\
\midrule
\midrule
\multirow{6}{*}{Color Jitter} & Prob    &  0.8  & 0.8  & 0.8   \\
\cmidrule{2-5}
& Brightness    &  0.4   &  0.4   & 0.4  \\
\cmidrule{2-5}
& Contrast  &  0.4  &  0.4   &  0.4 \\
\cmidrule{2-5}
& Saturation  &   0.2  & 0.2  & 0.2   \\
\cmidrule{2-5}
& Hue  &  0.1    & 0.1  & 0.1   \\
\midrule
\midrule
\multirow{1}{*}{Gray Scale} & Prob     &  0.2   &  0.2  &  0.2    \\
\midrule
\midrule
\multirow{1}{*}{Gaussian Blur} & Prob    & 1.0 / 0.1 & 1.0 / 0.1 & 1.0 / 0.1 \\
\midrule
\midrule
\multirow{1}{*}{Solarization} & Prob    & 0.0 / 0.2 & 0.0 / 0.2 & 0.0 / 0.2 \\
\midrule
\midrule
\multirow{1}{*}{Horizontal Flip} & Prob  & 0.5  & 0.5  & 0.5  \\
\bottomrule
\end{tabular}
\caption{The augmentation parameter values for BYOL, SimSiam, and Barlow Twins. Note that some entries are asymmetric across the two views. Such numbers are shown as A/B.}
\label{tab:others_aug}
\end{table*}

\subsection{SSL methods training}
While we follow the default settings for MoCo, SimSiam, and Barlow Twins provided in the solo-learn library~\citep{da2022solo}, we increase the batch size and decrease the learning rate parameters in SimCLR and BYOL based on their original papers~\citep{chen2020simple,grill2020bootstrap}, as this yields better performance compared to the solo-learn settings. We also use the learning rate scaling rule: $lr = base\_lr \times batch\_size/256$. We present the training details in Tables~\ref{tab:train_config1} and \ref{tab:train_config2}.

\begin{table*}[hbt!]
\centering
\begin{tabular}{l | c | c | c}
\toprule
 \textbf{Config} & \textbf{SimCLR} & \textbf{MoCo} & \textbf{BYOL} \\
\bottomrule
Optimizer & LARS & SGD & LARS \\
Base lr & 0.3 & 0.3 & 0.2 \\
Weight decay & 1e-6 & 3e-5 & 15e-7\\
Batch size & 256 & 64 & 256 \\
Learning rate schedule & Cosine decay & Cosine decay & Cosine decay \\
Warmup epochs & 10 & 10 & 10 \\
Backbone & Resnet50 & Resnet50 & Resnet50  \\
Base momentum backbone & NA & 0.99 & 0.99\\
Final momentum backbone & NA &  0.999  & 1.0\\
Projector hidden dimension & 4096 & 2048  & 4096\\
Projector output dimension & 512 & 256  & 256 \\
Predictor hidden dimension & NA & NA  & 4096 \\
Queue size & NA & 65536 & NA \\
Temprature for contrastive loss & 0.2 &  0.2 & NA \\
\end{tabular}
\caption{SSL pretraining setting for SimCLR, MoCo and BYOL.}
\label{tab:train_config1}
\end{table*}

\begin{table*}[hbt!]
\centering
\begin{tabular}{l | c | c }
\toprule
 \textbf{Config} & \textbf{SimSiam} & \textbf{Barlow Twins}  \\
\bottomrule
Optimizer & SGD & LARS \\
Base lr & 0.5 & 0.8 \\
Weight decay & 1e-5 & 1.5e-6\\
Batch size & 64 & 64  \\
Learning rate schedule & Cosine decay & Cosine decay  \\
Warmup epochs & 10 & 10  \\
Backbone & Resnet50 & Resnet50  \\
Projector hidden dimension & 4096 & 4096  \\
Projector output dimension & 4096 & 4096  \\
Predictor hidden dimension & 512 & NA   \\
Cross-covariance matrix off-diagonal scaling factor & NA & 0.0051  \\
Loss scaling factor & NA &  0.048  \\
\end{tabular}
\caption{SSL pretraining setting for SimSiam and Barlow Twins.}
\label{tab:train_config2}
\end{table*}

\subsection{Linear probing}
For evaluation, we adhere to the linear probing protocol of prior studies, which involves training a linear classifier on the output of the frozen encoder for both in-distribution and out-of-distribution downstream tasks. 
Since the LARS optimizer \citep{you2017large} has been shown to yield better results \citep{chen2021exploring}, we have chosen to use it as our optimizer. Additionally, it is generally observed that regularization, such as weight decay, can hurt performance. Therefore, we have set the weight decay to 0. We present the linear probing details in Table~\ref{tab:linear_config}.

\begin{table*}[hbt!]
\centering
\begin{tabular}{l | c  }
\toprule
 \textbf{Config} & \textbf{Linear Probing}  \\
\bottomrule
Optimizer & LARS \\
Base lr &  0.1 \\
Weight decay & 0.0 \\
Batch size & 512  \\
Learning rate schedule & Cosine decay  \\
Warmup epochs &  0  \\
\end{tabular}
\caption{Linear probing setting for downstream evaluation.}
\label{tab:linear_config}
\end{table*}

\section{Complimentary results}

In this section, we provide complimentary results for the experiments presented in Section~\ref{sec:results} of our manuscript. Tables~\ref{tab:indist_top5} and \ref{tab:ood_top5} show the Top-1 and Top-5 accuracy results for experiments related to \textbf{RQ1}. Table \ref{tab:results_rq2} present the Top-1 and Top-5 accuracy results for experiments associated with \textbf{RQ2}. Table~\ref{tab:results_rq4} shows the Top-1 and Top-5 accuracy results for experiments related to \textbf{RQ4}.
\begin{table*}[hbt!]
\centering
\begin{tabular}{c | c  |  c |  c | c | c }
\toprule
 & \textbf{SimCLR} & \textbf{MoCo} & \textbf{BYOL} &  \textbf{SimSiam} & \textbf{Barlow Twins} \\
\midrule
Baseline & 61.84 / 84.00 & 60.23 / 82.90 & 62.20 / 84.04 & 65.52 / 86.88 & 67.74 / 87.99  \\
\midrule
ICGAN & 64.00 / 85.93 & 65.48 / 87.41 & 66.03 / 87.25 & 68.07 / 88.11 & 67.20 / 87.54  \\
\midrule
Stable Diff & 68.12 / 88.76 & 70.17 / 90.34 & 70.21 / 90.07 & 69.43 / 89.36 & 70.89 / 90.06  \\
\bottomrule
\end{tabular}
\caption{Top-1/Top-5 accuracy ($\%$) of linear probing for the baseline (only the standard augmentations) and the generative augmentation on ImageNet.}
\label{tab:indist_top5}
\end{table*}

\begin{table*}[hbt!]
\centering
\begin{tabular}{c|c|c|c|c|c|c}
\toprule
\multicolumn{1}{c}{} & & \textbf{Food101} & \textbf{CIFAR10} &\textbf{CIFAR100} & \textbf{Places365} & \textbf{iNaturalist}\\
\midrule
\multirow{4}{*}{\begin{turn}{-270}\textbf{SimCLR}\end{turn}} & Baseline & 58.03 / 83.17 & 61.74 / 95.73 &  36.88 / 66.92 & 45.74 / 77.32 & 20.17 / 36.88 \\
\cmidrule{2-7}
& ICGAN & 57.62 / 82.68 & 61.92 / 96.02 & 38.74 / 68.56 & 47.41 / 78.25 & 20.01 / 37.67\\
\cmidrule{2-7}
& Stable Diff & 62.76 / 86.44 & 63.35 / 95.99 & 40.17 / 70.32 & 48.93 / 79.87 & 24.13 / 44.04\\
\midrule
\midrule
\multirow{4}{*}{\begin{turn}{-270}\textbf{MoCo}\end{turn}} & Baseline & 54.50 / 80.21 & 58.17 / 95.43 & 34.28 / 64.74 & 45.70 / 76.62 & 16.52 / 31.70 \\
\cmidrule{2-7}
& ICGAN & 59.54 / 83.77 & 63.70 / 96.36 & 39.90 / 70.52 & 48.82 / 79.49 & 20.99 / 39.11\\
\cmidrule{2-7}
& Stable Diff & 63.97 / 87.40 & 65.14 / 96.66 & 41.21 / 72.00 & 49.91 / 80.84 & 26.29 / 47.11\\
\midrule
\midrule
\multirow{4}{*}{\begin{turn}{-270}\textbf{BYOL}\end{turn}} & Baseline & 54.14 / 80.07 & 55.34 / 94.26 &  28.70 / 57.82 & 46.93 / 77.57 & 6.91 / 21.68 \\
\cmidrule{2-7}
& ICGAN & 53.12 / 79.38 & 58.76 / 95.45 & 34.49 / 64.99 & 47.24 / 78.08 & 7.36 / 16.68\\
\cmidrule{2-7}
& Stable Diff & 58.85 / 84.01 & 61.90 / 96.28	& 38.71 / 68.75 & 49.09 / 80.12 & 10.74 / 23.64 \\
\midrule
\midrule
\multirow{4}{*}{\begin{turn}{-270}\textbf{SimSiam}\end{turn}} & Baseline & 60.71 / 84.65 & 61.93 / 96.08 & 37.69 / 67.89	& 48.78 / 79.64 & 22.07 / 40.46 \\
\cmidrule{2-7}
& ICGAN & 65.15 / 87.42 & 64.28 / 96.55 & 40.21 / 70.62	& 49.38 / 80.04 & 27.07 / 46.76\\
\cmidrule{2-7}
& Stable Diff & 64.85 / 87.32 & 63.88 / 96.41	& 40.58 / 70.86 & 49.66 / 80.43 & 28.25 / 48.64\\
\midrule
\midrule
\multirow{4}{*}{\begin{turn}{-270}\textbf{\shortstack{Barlow \\ Twins}}\end{turn}} & Baseline & 66.71 /  88.54	& 65.49 /  96.76 & 41.19 /  71.34	& 49.47 /  80.55 & 27.99 /  47.51 \\
\cmidrule{2-7}
& ICGAN  & 65.50 /  87.62	& 63.34 /  96.50	& 41.95 /  71.58	& 48.64 /  79.76	& 26.19 /  45.51\\
\cmidrule{2-7}
& Stable Diff & 69.97 /  90.37	& 65.20 /  96.70 & 43.47 /  72.93 & 50.31 /  81.22 & 32.42 /  53.31\\
\end{tabular}
\caption{Top-1/Top-5 accuracy ($\%$) results of linear evaluation on several datasets with different augmentation strategies.}
\label{tab:ood_top5}
\end{table*}

\begin{table*}[hbt!]
\centering
\begin{tabular}{c| c | c | c}
\toprule
\multicolumn{2}{c|}{\textbf{Probability}} & \multicolumn{2}{c}{\textbf{Imagenet}} \\
\multicolumn{2}{c|}{} & One View & Both Views \\
\midrule
\multirow{6}{*}{\begin{turn}{-270}\textbf{SimCLR}\end{turn}} &  0 & 61.84 / 84.00 & 61.84 / 84.00\\
\cmidrule{2-4}
& 0.25  & 66.01 / 88.42 & 67.26 / 88.11\\
\cmidrule{2-4}
& 0.5 & 67.62 / 88.92 & 68.12 / 88.76\\
\cmidrule{2-4}
& 0.75 & 68.11 / 89.01 & 67.34 / 88.35 \\
\cmidrule{2-4}
& 1.0 & 68.11 / 88.86 & 62.11 / 84.41 \\
\bottomrule
\end{tabular}
\caption{Top-1/Top-5 accuracy ($\%$) of different probability values of applying the generative augmentation to one view or both views, measured on the ImageNet validation set.}
\label{tab:results_rq2}
\end{table*}

\begin{table*}[!hbt]
\centering
\begin{tabular}{c| l | c | c | c}
\toprule
\multicolumn{2}{l|}{\textbf{Augmentation Strategy}}  & \multicolumn{3}{c}{\textbf{Imagenet}}  \\

\multicolumn{2}{c|}{} & One View & Both Views & Both Views\\

\multicolumn{2}{c|}{} & $p = 1$ & $p = 0.5$ &  $p = 1$ \\
\midrule
\multirow{6}{*}{\begin{turn}{-270}\textbf{SimCLR}\end{turn}} & Only standard &  \multicolumn{3}{c} {61.84 / 84.00}\\
\cmidrule{2-5}
& Only Generative  & 63.80 / 86.00  & 65.40 / 87.12   & 57.83 / 81.69\\
\cmidrule{2-5}
& Generative \& Random Crop & 67.15 / 88.22  & 66.98 / 88.08  & 58.49 / 81.87\\
\cmidrule{2-5}
& Generative \& standard & 68.11 / 88.86  & 68.12 / 88.76  & 62.11 / 84.41\\
\bottomrule
\end{tabular}
\caption{Top-1/Top-5 accuracy ($\%$) of applying different augmentation strategies with three levels of applying generative augmentation, measured on the ImageNet validation set. The variable $p$ denotes the probability of applying generative augmentation, while "one/both views" refers to whether generative augmentation is applied to one or both views of the data.}
\label{tab:results_rq4}
\end{table*} 

\section{Generative augmentation samples}
\begin{figure*}[ht]
    \centering
    \includegraphics[width=0.75\textwidth]{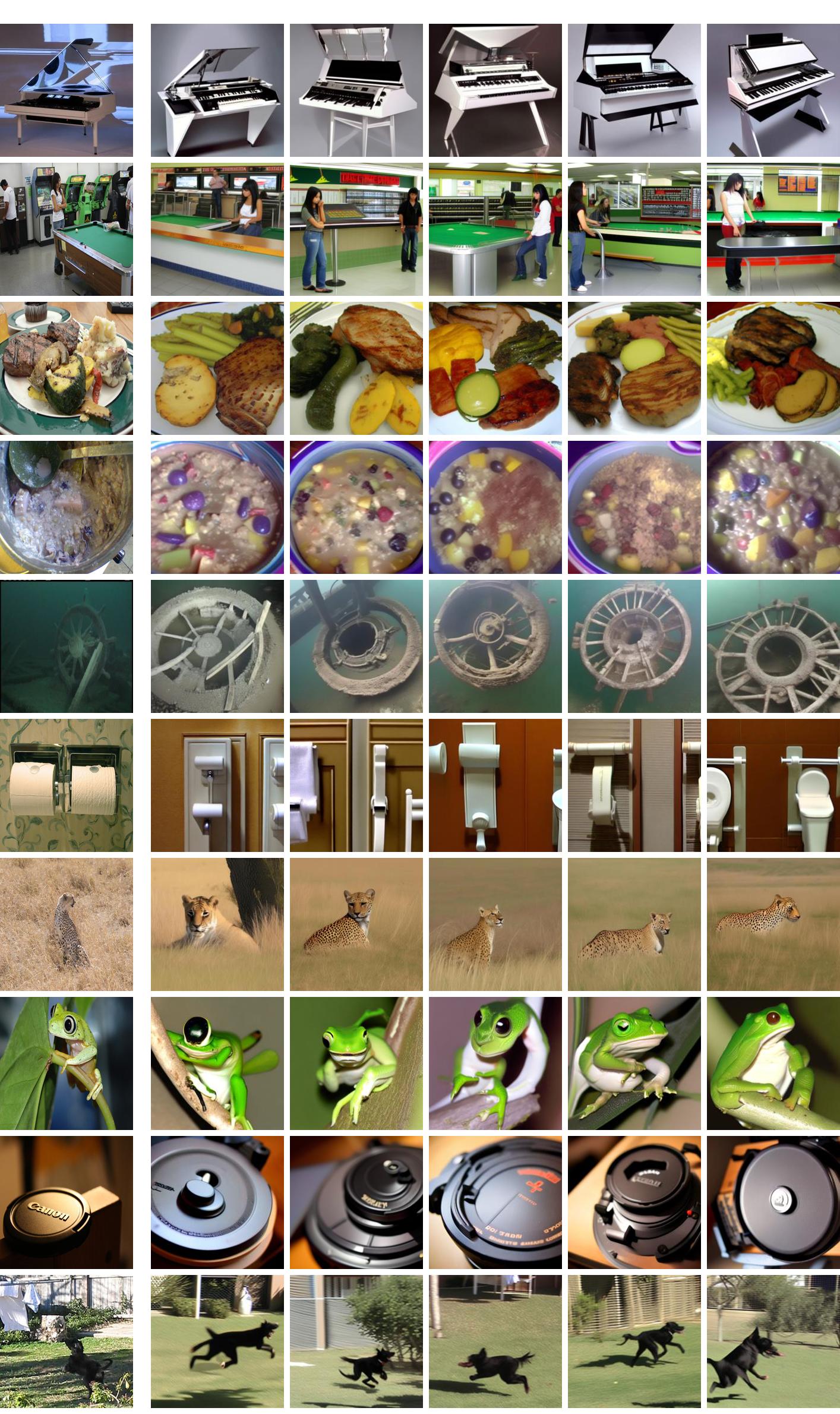}
    \caption{Augmentations generated by Stable Diffusion. The first column shows the original images, and the other columns show the generated augmentations.}
    \label{fig:stable_diff}
\end{figure*}

\begin{figure*}[ht]
    \centering
    \includegraphics[width=0.75\textwidth]{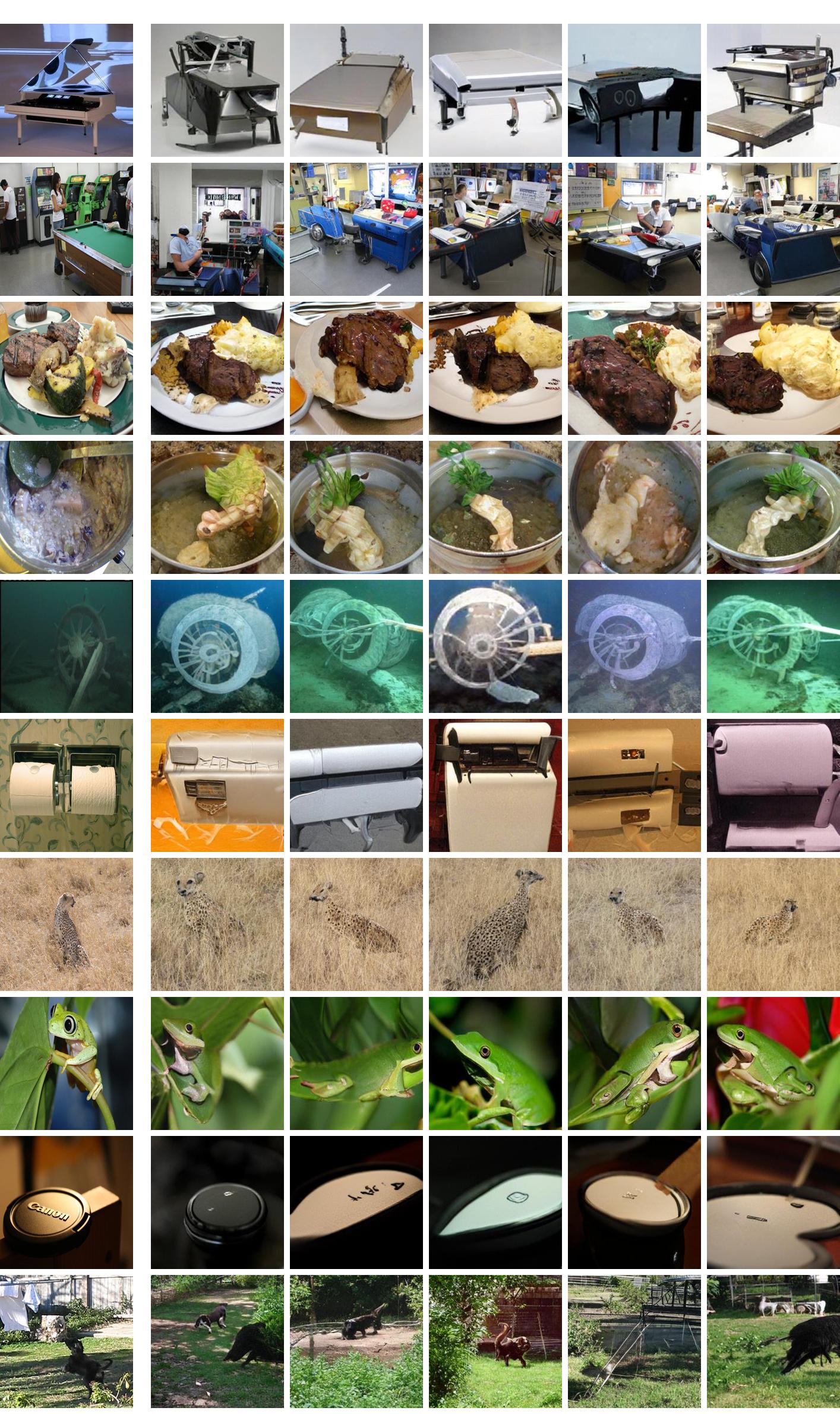}
    \caption{Augmentations generated by ICGAN. The first column shows the original images, and the other columns show the generated augmentations.}
    \label{fig:icgan}
\end{figure*}

Fig.~\ref{fig:stable_diff} illustrates augmentations generated using Stable Diffusion, and Fig.~\ref{fig:icgan} shows augmentations generated using ICGAN. Stable Diffusion is trained on image-text data using classifier-free guidance. In our approach, we condition it solely on the representation of the input image to generate various augmentations. On the other hand, ICGAN is a conditional GAN where the goal of its generator is to produce realistic images similar to the neighbors of each given instance. In ICGAN, images in the same neighborhood may belong to different classes. Therefore, we expect to see instances that may not be from the same class but carry a close representation to the input image.

\section{Computation}
\textbf{Offline generation:} Since generating augmentations on the fly would be time-consuming, we pre-generated augmented instances for the training split of ImageNet, which consists of approximately 1.2 million images. 
Using Stable Diffusion and ICGAN separately, we ran 10 parallel processes for each model on A100 GPUs, with each process generating one augmented sample per image (resulting in 10 samples in total for each method). Using Stable unCLIP source code\footnote{\url{https://huggingface.co/docs/diffusers/en/api/pipelines/stable_unclip}}, each round of generation over the entire ImageNet took $\thicksim$178 hours. In contrast, ICGAN was four times faster, taking $\thicksim$40 hours to iterate over the entire ImageNet dataset.\\
\textbf{SSL pretraining:} Since we pre-generated augmented samples, during training, we randomly selected one of the 10 samples for each image and applied the appropriate standard augmentation. This process allowed our training time to be equivalent to any other SSL training. We pretrained each SSL method on 4 A40 GPUs, with each run taking $\thicksim$34 hours. It is worth noting that loading ImageNet was our main bottleneck. By using the DALI\footnote{\url{https://docs.nvidia.com/deeplearning/dali/user-guide/docs/examples/frameworks/pytorch/pytorch-basic_example.html}} dataloader, we reduced the training time by almost half.\\
\textbf{Linear Probing:} Computation time for linear probing is highly dependent on the size of the dataset. Using 4 T4v2 GPUs, it took  $\thicksim$14 hours for ImageNet, $\thicksim$10 minutes for CIFAR10,  $\thicksim$10 minutes for CIFAR100, $\thicksim$10 hours for Places365 and $\thicksim$10 hours for iNaturalist.

\end{document}